%% file: main.tex
\documentclass[]{spie}

\usepackage{amsmath,amsfonts,amssymb}
\usepackage{graphicx}
\usepackage[colorlinks=true, allcolors=blue]{hyperref}
\usepackage{booktabs}
\usepackage{multirow}
\usepackage{threeparttable}
\usepackage{rotating}
\usepackage{subcaption}

\usepackage{caption}

\title{Exploring optical flow inclusion into nnU-Net framework for surgical instrument segmentation}

\author[a,b,c]{Marcos Fernández-Rodríguez}
\author[a,b,c]{Bruno Silva}
\author[a,b]{Sandro Queirós}
\author[c,d]{Helena R. Torres}
\author[c,d,e]{Bruno Oliveira}
\author[c,d]{Pedro Morais}
\author[f]{Lukas R. Buschle}
\author[a,b]{Jorge Correia-Pinto}
\author[a,b]{Estevão Lima}
\author[c,d]{João L. Vilaça}
\affil[a]{Life and Health Sciences Research Institute (ICVS), School of Medicine, University of Minho, Braga, Portugal.}
\affil[b]{ICVS/3B’s - PT Government Associate Laboratory, Braga/Guimarães, Portugal.}
\affil[c]{2Ai – School of Technology, IPCA, Barcelos, Portugal.}
\affil[d]{LASI – Associate Laboratory of Intelligent Systems, Guimarães, Portugal.}
\affil[e]{Algoritmi Center, School of Engineering, University of Minho, Guimarães, Portugal.}
\affil[f]{KARL STORZ SE \& Co. KG, Tuttlingen, Germany.}

\authorinfo{Further author information: (Send correspondence to M. Fernández-Rodríguez)\\ Authors emails:\\ Marcos Fernández-Rodríguez: id9877@alunos.uminho.pt; Bruno Silva: id9875@alunos.uminho.pt; Sandro Queirós: sandroqueiros@med.uminho.pt; Helena R. Torres: htorres@ipca.pt; Bruno Oliveira: boliveira@ipca.pt; Pedro Morais: pmorais@ipca.pt; Lukas R. Buschle: Lukas.buschle@karlstorz.com; Jorge Correia-Pinto: jcp@med.uminho.pt; Estevão Lima: estevaolima@med.uminho.pt; João L. Vilaça: jvilaca@ipca.pt}

\begin{document} 
\maketitle

\begin{abstract}
Surgical instrument segmentation in laparoscopy is essential for computer-assisted surgical systems. Despite the Deep Learning progress in recent years, the dynamic setting of laparoscopic surgery still presents challenges for precise segmentation. The nnU-Net framework excelled in semantic segmentation analyzing single frames without temporal information. The framework's ease of use, including its ability to be automatically configured, and its low expertise requirements, have made it a popular base framework for comparisons. Optical flow (OF) is a tool commonly used in video tasks to estimate motion and represent it in a single frame, containing temporal information. This work seeks to employ OF maps as an additional input to the nnU-Net architecture to improve its performance in the surgical instrument segmentation task, taking advantage of the fact that instruments are the main moving objects in the surgical field. With this new input, the temporal component would be indirectly added without modifying the architecture. Using CholecSeg8k dataset, three different representations of movement were estimated and used as new inputs, comparing them with a baseline model. Results showed that the use of OF maps improves the detection of classes with high movement, even when these are scarce in the dataset. To further improve performance, future work may focus on implementing other OF-preserving augmentations.
\end{abstract}


\section{INTRODUCTION}
\label{sec:intro} 

Surgical instruments segmentation in laparoscopy is an essential base for computer-assisted surgical systems to develop diverse applications, which would involve tracking, phase recognition or pose estimation, among others. Despite recent efforts in the Deep Learning field through new architectures and different algorithmic approaches, the dynamic setting of laparoscopic surgery still makes it hard to obtain a precise segmentation\cite{lee2019segmentation}.

Since the apparition of the nnU-Net framework, a self-configuring method for deep learning single frame image segmentation, in its 2020 paper by Isensee et al.\cite{isensee2021nnu} demonstrating its potential and reaching state-of-the-art performance on 33 of 53 different international challenges of 2019, several works have tried to extend nnU-Net by modifying its architecture to enhance its performance and extending its capabilities\cite{baumgartner2021nndetection,isensee2023extending,mcconnell2022integrating,zhou2023nnformer}.

The ability to be automatically configured, taking into account features of the dataset, achieving state-of-the-art results, along with the fact that the framework was conceptualized to be run in graphic processing units (GPUs) with 11 Giga Bytes (GB) of random access memory (RAM) and the requirement of low expertise, helped nnU-Net to be established as a base framework for comparisons\cite{isensee2021nnu}. Trying to preserve this advantage in terms of accessibility, this work seeks to improve nnU-Net results in semantic surgical instrument segmentation by just including additional inputs to the framework with temporal data, differently from past works\cite{mcconnell2022integrating,zhou2023nnformer}, with minor changes in the automated pipeline and minor modifications in the original architecture.

Optical flow (OF) estimates motion and represents it in a single frame. OF has been widely used in video tasks and automated driving as a tool to extract motion features and depth\cite{liu2020learning,lai2019bridging}. The motion representation of an image has the potential of integrating temporal information, which may refine the semantic segmentation results\cite{rashed2019optical}. In the surgical domain, instruments are often what moves the most, and consequently, semantic surgical instrument segmentation is an excellent candidate application to evaluate this hypothesis.

The aim of this work is to study the OF influence in the framework provided by nnU-Net, and different methods to include it in the pipeline offered by the framework, while preserving its architecture. Moreover, this work intends to explore how including OF temporal information affects the performance of the surgical instrument segmentation task.

\section{METHODS}
\label{sec:methods} 
This work aims to explore the effects of OF information as an additional input for the nnU-Net, its impact in the results and potential to leverage semantic surgical instrument segmentation with minimal changes in its architecture. OF was estimated by comparing movement of pixels between frames of a video or image sequence, using the assumption that nearby pixels have similar motion. Two different values were chosen arbitrarily to obtain our estimations: (1) comparing the current frame with the previous frame (t1); and (2) comparing the current frame with the fifth last frame (t5). Three OF representations were prepared: a RGB image from the color wheel representation of movement (RGBof), X axis and Y axis displacement maps (XY), and polar representation in magnitude and angle (PC)\cite{rashed2019optical}.

Seven models were trained with nnU-Net, including standalone RGB images of the original dataset (RGB) as a baseline for comparison, and six targeting the combination of RGB images with one of the modalities explained above. To mitigate the native randomness of the framework, each model was trained four times and the results averaged. In short, the input variants were:
\renewcommand{\labelitemii}{\textperiodcentered}
\begin{itemize}
    \item RGB
    \item RGB with t1 or t5 OF:
    \begin{itemize}
        \item RGB + RGBof
        \item RGB + XY
        \item RGB + PC
    \end{itemize}
\end{itemize}

\begin{figure}[htbp]
    \begin{center}
    \begin{tabular}{c}
    \includegraphics[width=0.98\textwidth]{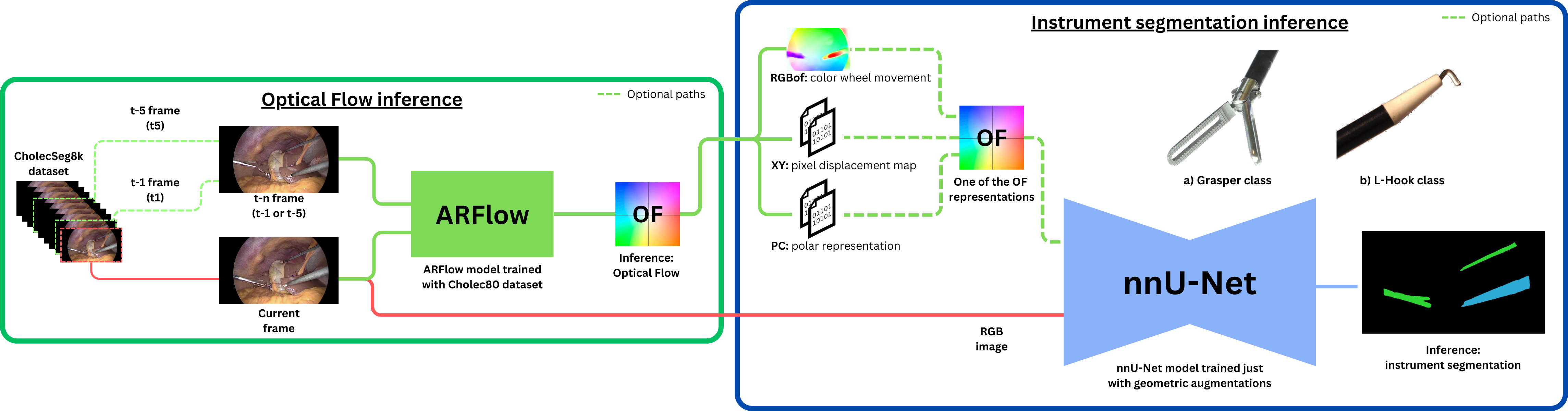}    
    \end{tabular}
    \end{center}
    \caption{Overview of the proposed workflow and segmentation classes: a) Grasper; b) L-hook.}
    \label{fig:workflow}
\end{figure}

One of the possible limitations is that OF cannot differentiate whether what is being moved is an instrument or a tissue being dragged. Consequently, introducing the OF limits the augmentations that can be applied to the input, requiring a deeper dive into the nnU-Net framework to implement more significant changes. Therefore, aiming to maintain accessibility, only geometric augmentations were considered for all trainings.

\subsection{Datasets}
For this purpose, two public datasets were used: (1) the Cholec80\cite{twinanda2016endonet}, employed to train ARFlow\cite{liu2020learning} and obtain the Optical Flow model; and (2) the CholecSeg8k\cite{hong2020cholecseg8k}, as the main dataset for surgical instrument segmentation evaluation.

\subsubsection{Cholec80}
Cholec80 consists of a dataset of laparoscopic cholecystectomies containing 80 videos captured at 25 frames per second (fps), which were downsampled to 1 fps for processing. The dataset is labeled with annotations about the surgical phase and instrument presence. Most videos have a 854 x 480 resolution, although a few have as much as 1920 x 1080 pixels\cite{twinanda2016endonet}.

\subsubsection{CholecSeg8k}
CholecSeg8k dataset is derived from Cholec80 and contains 8080 frames of 101 clips (80 frames each, captured at 25 fps) of 17 different laparoscopic cholecystectomy cases performed by 13 surgeons. The videos have a resolution of 854 x 480 pixels, and semantic segmentation annotations of 13 different classes (10 organs and tissues, and 2 instruments) are provided\cite{hong2020cholecseg8k}.

\begin{figure}[hptb]
    \begin{center}
    \begin{tabular}{c}
    \includegraphics[width=0.75\textwidth]{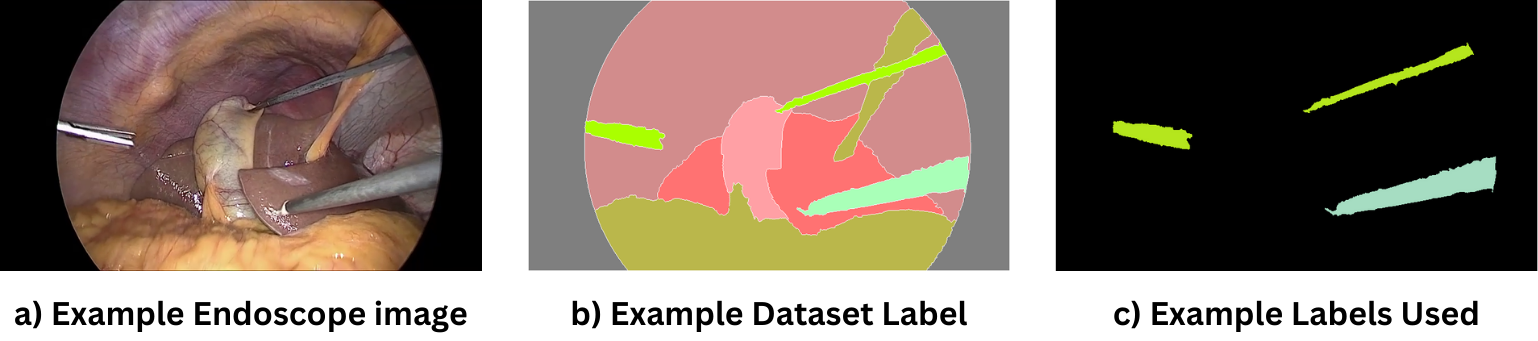}    
    \end{tabular}
    \end{center}
    \caption{Graphical representation of the dataset used.}
    \label{fig:dataset}
\end{figure}

Given the focus on surgical instrument segmentation, the dataset’s ground truth masks (Fig. \ref{fig:dataset}) were modified. Organs, tissues and background were assembled as “Background” class, while instruments remained as separate classes (“Grasper” and “L-hook”[Fig. \ref{fig:workflow} - a, b]), making a total of 3 classes. Table \ref{tab:tablei} shows the class distribution and the split used for training and test, having in mind that both instruments can appear in the same frame at the same time. This split was chosen to preserve dataset’s instrument class and frame balance (Table \ref{tab:tableii}), separating entire cases. Training set included cases 1, 12, 17, 25, 27, 28, 35, 43, and 48 ($\sim$58 \% total frames). Validation set included cases 18, 20, 24, and 37 ($\sim$22 \% total frames). Test set included cases 9, 26, 52, and 55 ($\sim$20 \% total frames).  Cross-validation was not used due to class unbalance constraints, which made impossible to generate four different groups which preserve class and frame balance.

\begin{table}[h]
\centering
\begin{threeparttable}
\caption{Frames and class balance per data subset}
\label{tab:tablei}
\begin{tabular}{lccc}
\toprule
                    & \textbf{Training set}    & \textbf{Validation set} & \textbf{Test set}   \\ 
\midrule
\textbf{Grasper frames}      & 3598 (76 \%)    & 1200 (68 \%)   & 1222 (76\%) \\
\textbf{L-hook frames}       & 1369 (29 \%)    & 426 (24 \%)    & 459 (29 \%) \\
\textbf{Total frames}        & 4720            & 1760           & 1600       \\
\midrule
\textbf{Dataset percentage\tnote{*}}  & 58.42 \%        & 21.78 \%       & 19.80 \%   \\
\bottomrule
\end{tabular}

\end{threeparttable}
\end{table}

\vspace{100pt}

\begin{table}[h]
\centering
\begin{threeparttable}
\caption{CholecSeg8k frames and class distribution by case and total}
\label{tab:tableii}
\begin{tabular}{ccccc}
\toprule
& \textbf{Case} & \textbf{Frames} & \textbf{Grasper} & \textbf{L-hook} \\
\midrule
\multirow{9}{*}{\textbf{Train}} 
& 1   & 1280 & 849 & 806 \\
& 12  & 640  & 577 & 558 \\
& 17  & 320  & 240 & 1   \\
& 25  & 320  & 320 & 0   \\
& 27  & 400  & 304 & 0   \\
& 28  & 560  & 326 & 0   \\
& 35  & 240  & 240 & 0   \\
& 43  & 720  & 502 & 0   \\
& 48  & 240  & 240 & 4   \\
\midrule
\multirow{4}{*}{\textbf{Validation}} 
& 18  & 160  & 160 & 160 \\
& 20  & 160  & 160 & 160 \\
& 24  & 960  & 400 & 0   \\
& 37  & 480  & 480 & 106 \\
\midrule
\multirow{5}{*}{\textbf{Test}} 
& 9   & 240  & 0   & 0   \\
& 26  & 320  & 182 & 0   \\
& 52  & 800  & 800 & 240 \\
& 55  & 240  & 240 & 219 \\
\midrule
&\textbf{Total} & 8080 & 6020 & 2254 \\
\bottomrule
\end{tabular}
\end{threeparttable}
\end{table}

\subsection{Optical flow}
The lack of ground truth OF maps for both datasets led to the use of an unsupervised method for their generation. ARFlow was selected as the method due to its low computational requirements and state-of-the-art performance\cite{liu2020learning}.

To obtain the optical flow, first, it was necessary to get a trained model of the ARFlow, which was obtained through the use of the Cholec80 dataset. Starting with the original resolution of the image, a random square crop was carried out with the size of the smaller side of the image. After that, the cropped image was rescaled to 256 x 256 pixels and fed to the network. The result was a trained model which outputs 256 x 256 OF fields of the respective input image. 

With this model, and in order to run inference on the CholecSeg8k dataset, images were cropped in two squares with the smaller image size, one aligned to the left of the image and the other on the right. These images were then rescaled to 256 x 256 pixels and the OF field inferred. At the end, both images were used to rebuild the original aspect ratio, with a 456 x 256 pixels resolution.

\subsection{nnU-Net}
A custom trainer was prepared for including optical flow in the nnU-Net pipeline, making use of the 2D version of U-net. Some modifications were required for the addition of the OF input, focused on performing exclusively geometrical augmentations, to preserve the OF information. Geometric augmentations included: rotations, scaling, mirroring and elastic deformations. In turn, Gaussian noise, Gaussian blur, changes in brightness, contrast, gamma and simulations of low resolution were removed from the original pipeline\cite{isensee2021nnu}.

Before being fed to the nnU-Net, the inputs were separated by their channels due to the framework treating them as different imaging modalities. In the case of OF arrays, rescaling to the dataset resolution (856 x 480) was necessary. For further information about the architecture used inside the nnU-Net and how the framework works, please refer to the original paper\cite{isensee2021nnu}.
    
\subsection{Evaluation metrics}
After training, the Dice coefficient (DC) was calculated on the test set (average plus standard deviation) for each instrument (Grasper DC and L-hook DC), to evaluate the ability to distinguish instruments from each other, and the average of both classes (Mean). Recall and precision were reported too.

\section{RESULTS}
Results are summarized in Table \ref{tab:tableiii}, containing the average of the results of the four trainings of each variant (full table with all results by training can be found in appendix A). Best results are highlighted in bold.

\begin{table}[bp]
\centering
\begin{threeparttable}
\caption{Performance Metrics}
\label{tab:tableiii}
\begin{tabular}{rccccccccc}
\toprule
    & \multicolumn{3}{c}{\textbf{Grasper}}  &       \multicolumn{3}{c}{\textbf{L-Hook}}     & \multicolumn{3}{c}{\textbf{Mean}} \\
      \cmidrule(lr){2-4} \cmidrule(lr){5-7} \cmidrule(lr){8-10}
      & \textbf{DC}             & \textbf{Recall}    & \textbf{Prec.} & \textbf{DC}             & \textbf{Recall}    & \textbf{Prec.} & \textbf{DC}             & \textbf{Recall}    & \textbf{Prec.} \\
\midrule
\textbf{RGB}         & 75.98\% & 72.82\% & 84.70\% & 31.97\% & 46.16\% & 29.16\% & 53.97\% & 59.49\% & 56.93\% \\
\textbf{t1 RGBof}    & 77.36\% & 73.07\% & \textbf{87.84}\% & 48.10\% & 68.79\% & 45.17\% & 62.73\% & 70.93\% & \textbf{66.50}\% \\
\textbf{t5 RGBof}    & 76.60\% & 72.93\% & 85.69\% & \textbf{49.68}\% & \textbf{69.94}\% & \textbf{46.10}\% & \textbf{63.14}\% & \textbf{71.43}\% & 65.89\% \\
\textbf{t1 XY}       & 77.10\% & 73.73\% & 86.30\% & 47.67\% & 66.55\% & 44.67\% & 62.39\% & 70.14\% & 65.48\% \\
\textbf{t5 XY}       & \textbf{77.77}\% & \textbf{74.27}\% & 87.29\% & 45.95\% & 66.67\% & 42.47\% & 61.86\% & 70.47\% & 64.88\% \\
\textbf{t1 PC}       & 76.94\% & 73.25\% & 86.08\% & 43.14\% & 62.23\% & 40.84\% & 60.04\% & 67.74\% & 63.46\% \\
\textbf{t5 PC}       & 76.52\% & 72.51\% & 86.02\% & 44.47\% & 67.08\% & 41.09\% & 60.49\% & 69.79\% & 63.55\% \\
\bottomrule
\end{tabular}
\begin{tablenotes}\footnotesize
    \item \textbf{Prec.}: Precision.
\end{tablenotes}
\end{threeparttable}
\end{table}

\subsection{Class performance analysis}
All comparisons are from OF variants against RGB version. In general, Mean DC was improved, achieving an average enhancement between all the OF variants of +7.80\%. Recall and precision increased, in average, by +10.59\% and +8.03\%, respectively. The variant with the best average result was the t5 RGBof for DC (+9.17\%) and recall (+11.94\%) and the t1 RGBof for precision (+9.57\%).

The Grasper class got just a mild improvement in DC, being +1.07\% the average of all the OF variants, as well as +0.47\% and +1.84\% for recall and precision. The variant with the best results for Grasper in DC and recall was the t5 XY, achieving +1.79\% and +1.45\% respectively, and the best precision was from the t1 RGBof variant with +3.14\%.  

Best enhancement is found in the L-hook class, with the best results observed for the RGBof variant with +17.71\%, +23.78\% and +16.94\% improvement for DC, recall and precision.

After analyzing each OF variant, no significant difference was observed between the distinct representations. In Mean, the bigger differences were 3.10\% in DC, 3.69\% in recall, and 3.04\% in precision. For Grasper, the bigger differences were 1.26\% in DC, 1.76\% in recall, and 2.15\% for precision. Finally, for the L-hook class, the bigger differences were 6.54\% for DC, 7.71\% for recall, and 5.26\% for precision.

\subsection{Groups analysis}
Table \ref{tab:tableiv} presents the results grouped by type of input (RGB baseline vs. OF inclusion) and by OF variant (t1 vs. t5, and each type of representation regardless of the time step).

\begin{table}[hbp]
\centering
\begin{threeparttable}
\caption{Grouped mean results by type of input (RGB vs. OF), temporal approach (t1 vs. t5), and OF representation (RGBof vs. XY vs. PC)}
\label{tab:tableiv}
\begin{tabular}{rccccccccc}
\toprule
                & \multicolumn{3}{c}{\textbf{Grasper}} & \multicolumn{3}{c}{\textbf{L-Hook}} & \multicolumn{3}{c}{\textbf{Mean}} \\
                \cmidrule(lr){2-4} \cmidrule(lr){5-7} \cmidrule(lr){8-10}
                & \textbf{DC} & \textbf{Recall} & \textbf{Precision} & \textbf{DC} & \textbf{Recall} & \textbf{Precision} & \textbf{DC} & \textbf{Recall} & \textbf{Precision} \\
\midrule
\textbf{RGB}        & 75.98\% & 72.82\% & 84.70\% & 31.97\% & 46.16\% & 29.16\% & 53.97\% & 59.49\% & 56.93\% \\
\textbf{OF}         & 77.05\% & 73.29\% & 86.53\% & 46.50\% & 66.88\% & 43.39\% & 61.77\% & 70.08\% & 64.96\% \\
\midrule
\textbf{t1}         & 77.14\% & 73.35\% & 86.74\% & 46.30\% & 65.86\% & 43.56\% & 61.72\% & 69.60\% & 65.15\% \\
\textbf{t5}         & 76.96\% & 73.23\% & 86.33\% & 46.70\% & 67.90\% & 43.22\% & 61.83\% & 70.57\% & 64.77\% \\
\midrule
\textbf{RGBof}      & 76.98\% & 73.00\% & 86.76\% & 48.89\% & 69.37\% & 45.64\% & 62.93\% & 71.18\% & 66.20\% \\
\textbf{XY}         & 77.44\% & 74.00\% & 86.79\% & 46.81\% & 66.61\% & 43.57\% & 62.12\% & 70.30\% & 65.18\% \\
\textbf{PC}         & 76.73\% & 72.88\% & 86.05\% & 43.80\% & 64.65\% & 40.96\% & 60.27\% & 68.77\% & 63.51\% \\
\bottomrule
\end{tabular}
\begin{tablenotes}\footnotesize
    \item \textbf{RGB}: contains standalone 'RGB' means; \textbf{OF}: contains 't1 RGBof', 't5 RGBof', 't1 XY', 't5 XY', 't1 PC', and 't5 PC' means; \textbf{t1}: contains 't1 RGBof, 't1 XY', and 't1 PC' means; \textbf{t5}: contains 't5 RGBof', 't5 XY', and 't5 PC' means; \textbf{RGBof}: contains 't1 RGBof', and 't5 RGBof' means; \textbf{XY}: contains 't1 XY', and 't5 XY' means; \textbf{PC}: contains 't1 PC', and 't5 PC' means.
\end{tablenotes}
\end{threeparttable}
\end{table}

Comparing t1 vs t5, there were almost no differences, being the most noticeable the 2.04\% improvement in L-hook recall for the t5 variants. The remaining metrics showed a variation lower than 1\%.

Regarding the different OF representations, the RGBof variant showed, in general, the best results, outperforming the other two variants for both L-hook class and Mean, but trailing the XY representation in the Grasper class. 
The largest performance drop was around 1\% against the XY representation in the Grasper class DC, recall and precision, but a 5\% improvement in the metrics was observed in the L-hook class, which led to an overall Mean improvement of nearly 2.5\% in DC, recall and precision.

\section{DISCUSSION}
At first glance, a clear improvement could be seen in the RGB baseline mean results due to the OF inclusion, independently of the OF variant used. Although this gain was small for the Grasper class, large performance increases were observed for the L-hook class. This class was the one with most movement and with the lowest amount of frames in the dataset, appearing in just 29\% of the images.

\begin{figure}[hptb]
    \centering
    \begin{subfigure}[c]{0.48\textwidth}
        \centering
        \includegraphics[width=\linewidth]{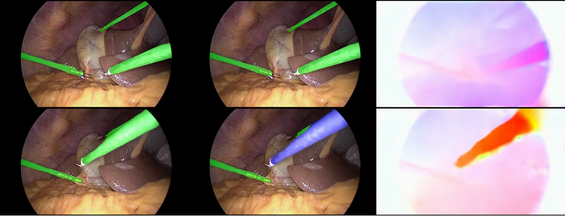}    
        \label{fig:ex1}
    \end{subfigure}
    \hfill
    \begin{subfigure}[c]{0.48\textwidth}
        \centering
        \includegraphics[width=\linewidth]{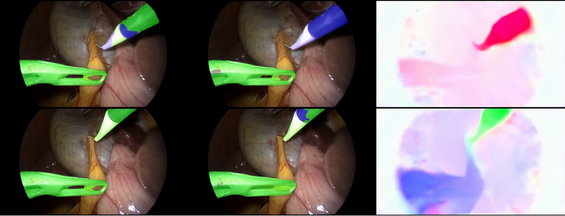}
        \label{fig:ex2}
    \end{subfigure}
    \caption{Four examples of the segmentation results for the 'RGB' (left) and 't1 RGBof' (middle) variants, plus the 'RGBof' image (right) used as OF input in the latter.}
    \label{fig:OF_examples}
\end{figure}

These results may suggest that the classes with most movement are the ones which may get the greater benefits from OF inclusion (Fig. \ref{fig:OF_examples}). Future work should however corroborate this finding quantitatively, by evaluating the performance difference according to the instruments' movement.

On the other hand, after looking for the different temporal approaches, a mild performance increase was found for the t5 variants. Nonetheless, these differences were not sufficiently significant to suggest a real improvement, namely considering the inherent fluctuation in scores due to the training randomness. Since only two time points were considered in our analysis, future work may consider extending this analysis to other time points in order to determine the one that provides the richest temporal representation for instrument segmentation. 

Lastly, looking towards the diverse representations used, results suggest that RGBof outperforms the others variants. This might be caused by the influence of the applied augmentations, which may affect less this representation, given its graphical nature, than those with the more numerical content (XY and PC). Regarding this point, more in deep research is required to adapt the part of the nnU-Net framework responsible for applying augmentations, to chose correctly which augmentations can be used and to adapt them properly to the input characteristics.
\subsection{Limitations}
This study suffered from some limitations in the augmentations which had to be shortened, as a result of our own constraints maintaining nnU-Net architecture, performing minimal changes to the framework, weakening their effectiveness.
Furthermore, the dataset class unbalance, would had a significant influence in results. 

Moreover, the dataset used, CholecSeg8k, was labeled using a semi-automated segmentation routine, which generated some label inconsistencies. During our analysis, some of the errors detected were: disappearing labels (Fig \ref{fig:sub1}), joining black borders as a class (Fig \ref{fig:sub2}), blurry label edges (Fig \ref{fig:sub3}), mixture of classes (Fig \ref{fig:sub4}), holes in surgical tools, an organ (Fig \ref{fig:sub5}) being labeled as a tool, and the trocar being present in both classes (Fig \ref{fig:sub6}). These flaws in the dataset inevitably affect the learning process of the model.

\newcommand{\tabsize}{0.45}

\begin{figure}[hptb]
    \centering
    \begin{subfigure}[c]{\tabsize\textwidth}
        \centering
        \includegraphics[width=\linewidth]{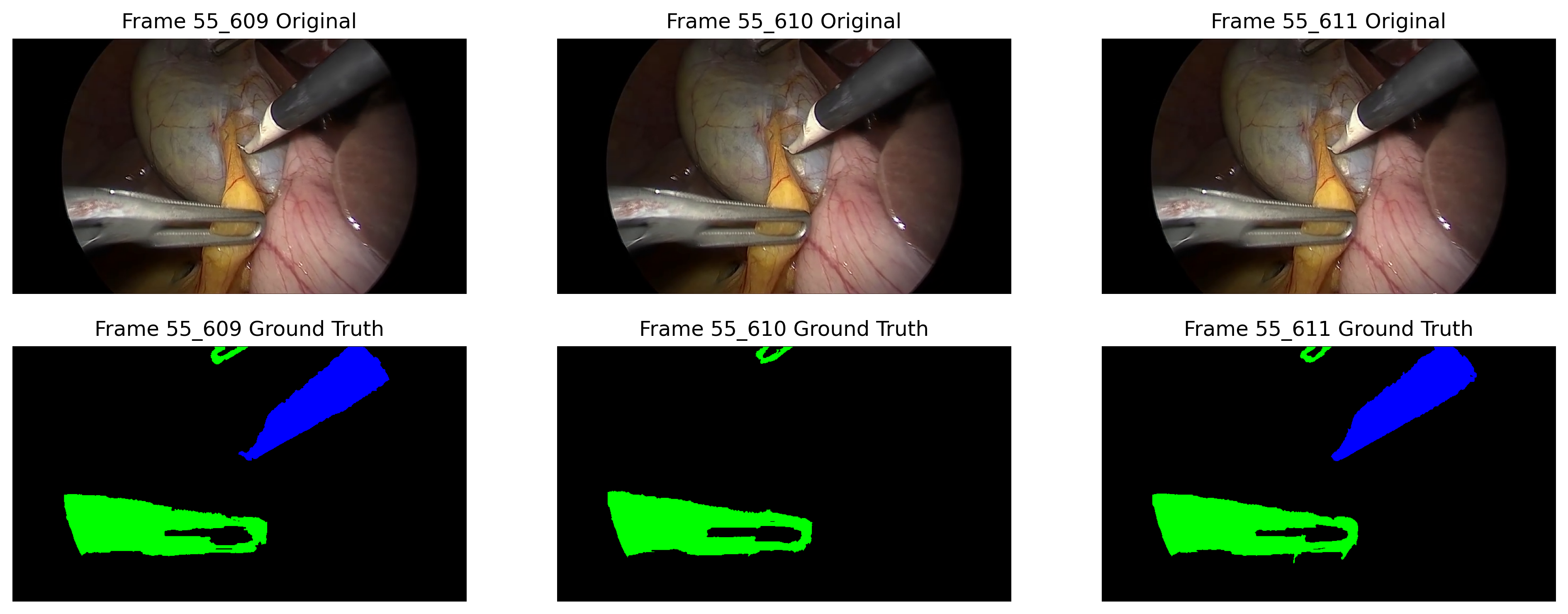}    
        \caption{Missing label}
        \label{fig:sub1}
    \end{subfigure}
    \hfill
    \begin{subfigure}[c]{\tabsize\textwidth}
        \centering
        \includegraphics[width=\linewidth]{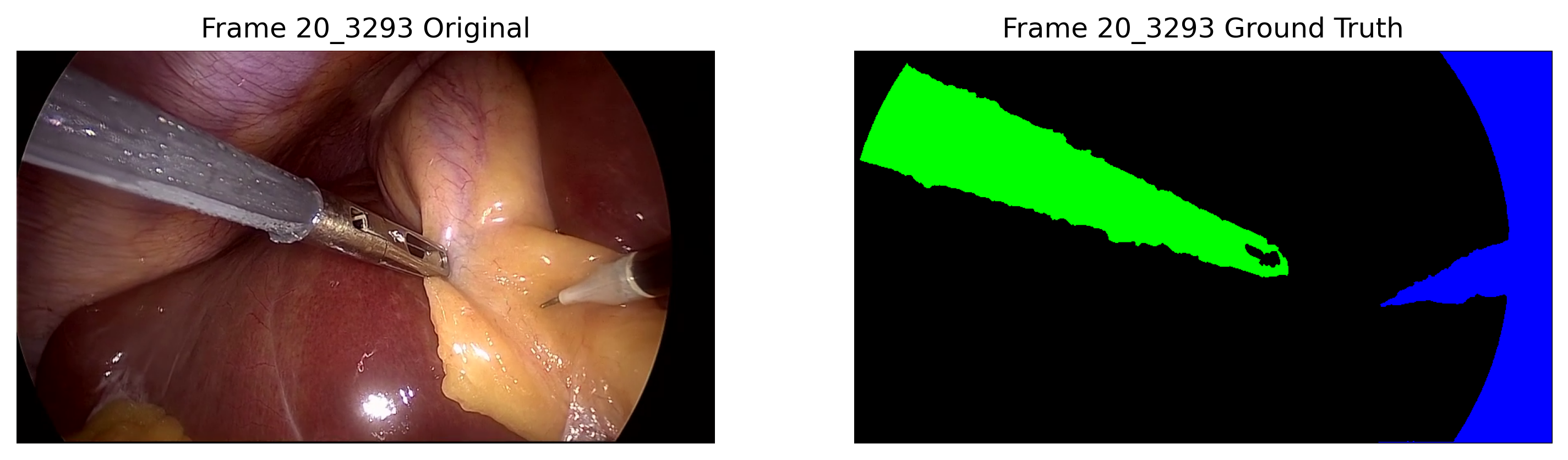}
        \caption{Border as a tool}
        \label{fig:sub2}
    \end{subfigure}

    \begin{subfigure}[c]{\tabsize\textwidth}
        \centering
        \includegraphics[width=\linewidth]{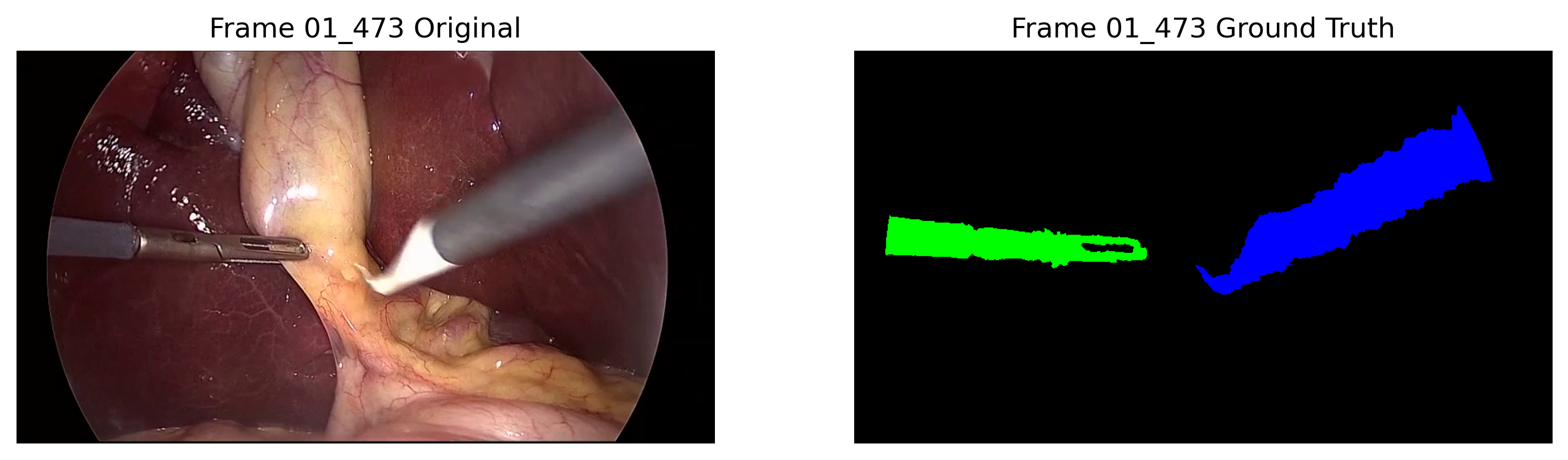}    
        \caption{Blurry label}
        \label{fig:sub3}
    \end{subfigure}
    \hfill
    \begin{subfigure}[c]{\tabsize\textwidth}
        \centering
        \includegraphics[width=\linewidth]{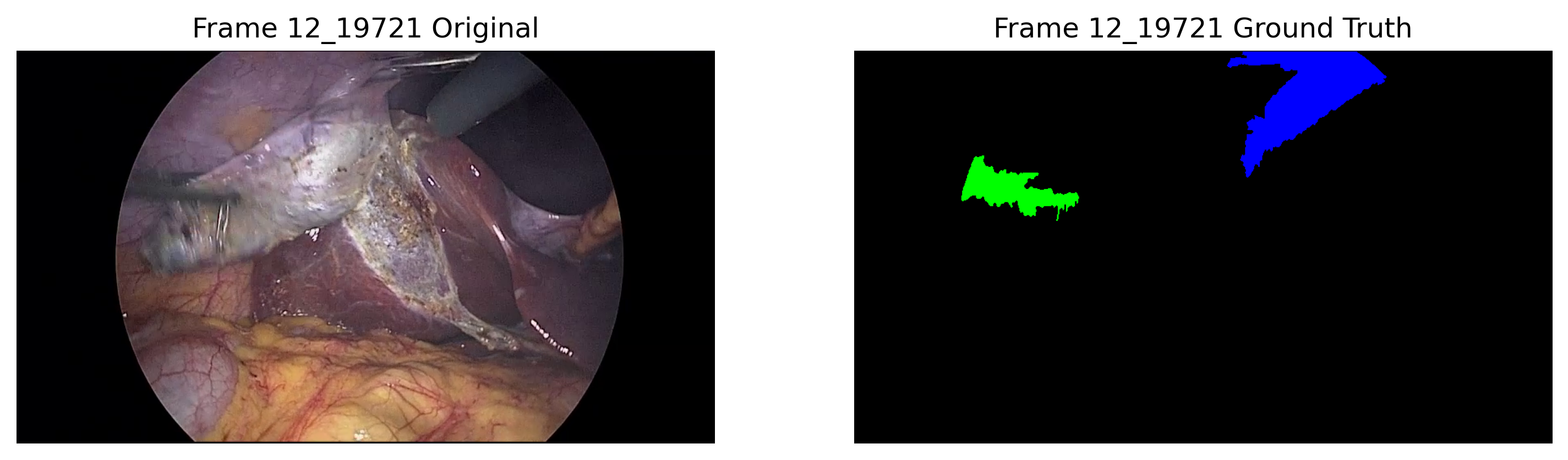}
        \caption{Grasper part labeled as L-hook}
        \label{fig:sub4}
    \end{subfigure}

    \begin{subfigure}[c]{\tabsize\textwidth}
        \centering
        \includegraphics[width=\linewidth]{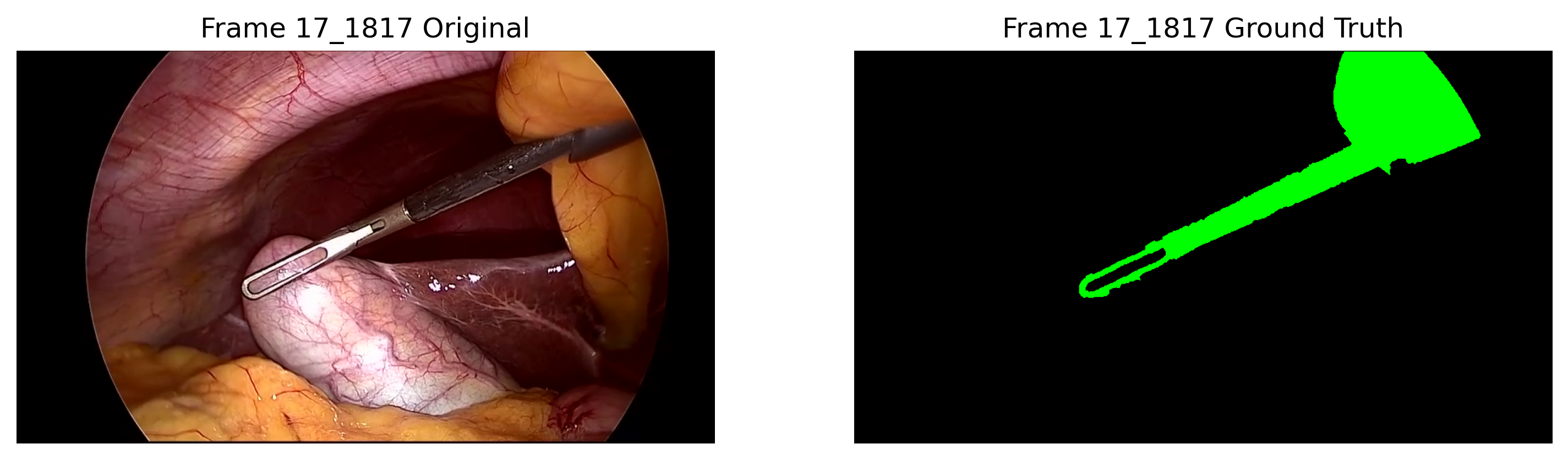}    
        \caption{Organ labeled as Grasper}
        \label{fig:sub5}
    \end{subfigure}
    \hfill
    \begin{subfigure}[c]{\tabsize\textwidth}
        \centering
        \includegraphics[width=\linewidth]{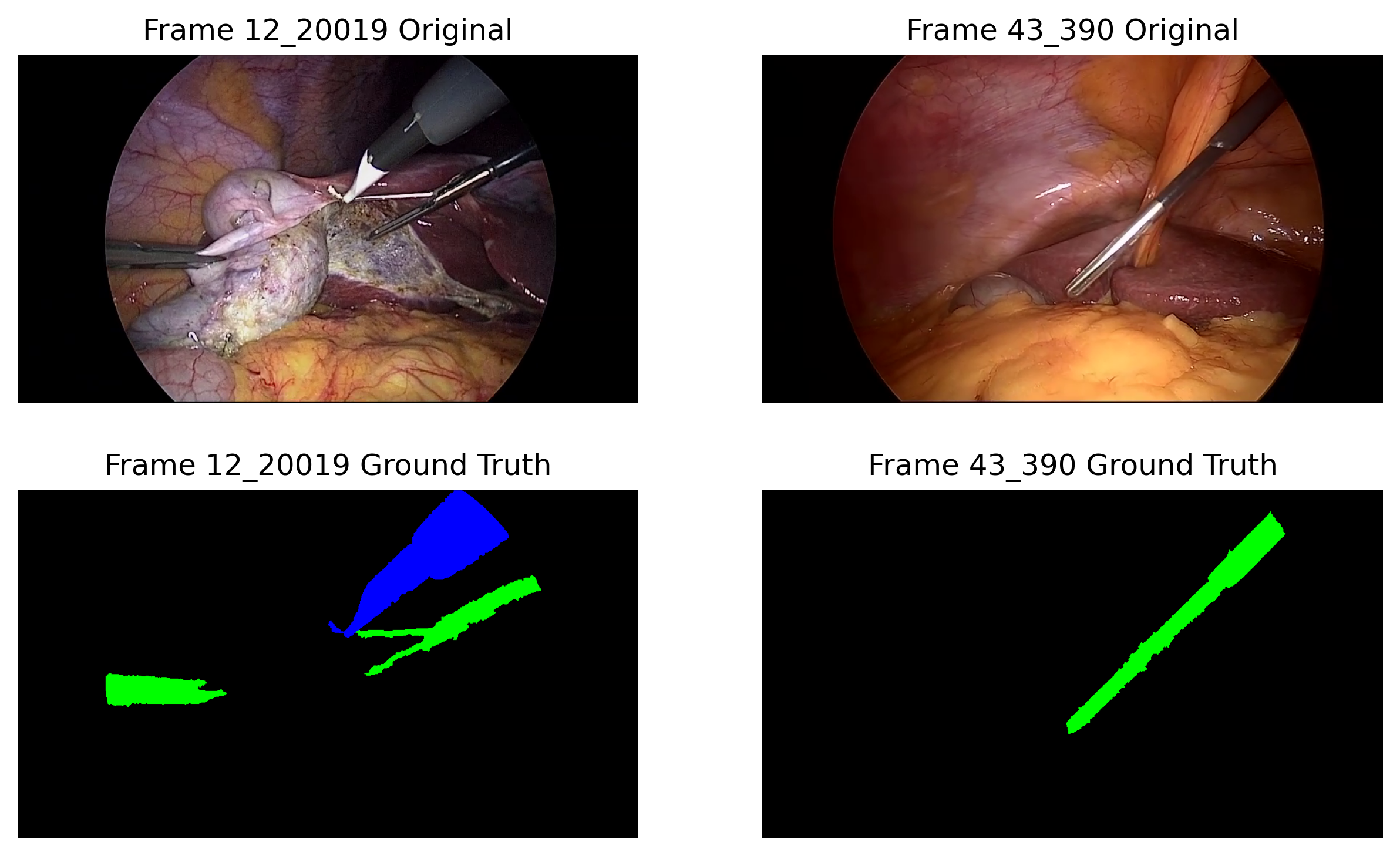}
        \caption{Trocar labeled as L-hook and Grasper}
        \label{fig:sub6}
    \end{subfigure}
    \caption{CholecSeg8k inconsistencies found during results analysis.}
\end{figure}

\newpage
\section{CONCLUSION}
The major contribution of this work is to demonstrate the ease of including OF in the nnU-Net architecture to improve results without performing big changes in the framework and keeping the advantages of low computational and expertise requirements.
Results demonstrated an improvement in L-hook DC, recall and precision, making the model more sensitive to the  class with less appearance and most movement in the dataset. These results must be taken carefully as some inconsistencies in the dataset were found during the results analysis.

Further research would be necessary to unravel how to extract all the potential of OF for semantic surgical  instrument segmentation, closely looking to how and where to apply augmentations, more carefully selecting which frame is used to compare and estimate OF, and considering a larger variety of datasets.

\acknowledgments          
 
The authors acknowledge the company KARL STORZ SE \& Co. KG, and the Portuguese National funds, through the Foundation for Science and Technology (FCT) - project UIDB/50026/2020 (DOI 10.54499/UIDB/50026/2020), UIDP~/50026/2020 (DOI 10.54499/UIDP/50026/2020), LA/P/0050/2020 (DOI 10.54499/LA/P/0050/2020), UIDB/05549/2020 (DOI: 10.54499/UIDB/05549/2020), UIDP/05549/2020 (DOI: 10.54499/UIDP/05549/2020), Lasi-LA/P/0104/2020, and grant CEECIND/03064/2018 (S.Q.; DOI 10.54499/CEECIND/03064/2018/CP1581\\/CT0017) for their support of this research.

\bibliography{report} 
\bibliographystyle{spiebib} 

\appendix    
\section{}
\label{sec:misc}
In this section, the complete data used to obtain Table \ref{tab:tableiii} in section 3 is shown.

\input{table_v2}

\end{document}

%% file: table_v2.tex
\begin{sideways}%
\setlength{\tabcolsep}{3pt}
\setlength{\belowcaptionskip}{2ex}
\begin{minipage}{1.17\textwidth}
\captionof{table}{Performance metrics across all trainings}
%\begin{tabular}{@{}rccccccccccc@{}}
\begin{tabular}{@{}cccccccccccc@{}}
\toprule
         & & \multicolumn{3}{c}{\textbf{Grasper}}                             & \multicolumn{3}{c}{\textbf{L-hook}}                               & \multicolumn{3}{c}{\textbf{Mean}}                             \\ 
          \cmidrule(lr){3-5} \cmidrule(lr){6-8} \cmidrule(lr){9-11}
         & \textbf{Train} & \textbf{DC}                            & \textbf{Recall}         & \textbf{Precision}       & \textbf{DC}                            & \textbf{Recall}         & \textbf{Precision}       & \textbf{DC}                            & \textbf{Recall}         & \textbf{Precision}       \\
\midrule
\multirow{4}{*}{\textbf{RGB}} 
     &   0                             & 75.97\% ± 0.267 & 72.82\%         & 84.75\%                       & 30.91\% ± 0.413 & 47.46\%         & 28.14\%                       & 53.44\% ± 0.340  & 60.14\%         & 56.45\%         \\
          & 1                             & 74.88\% ± 0.289 & 71.74\%         & 82.91\%                       & 36.18\% ± 0.431 & 49.61\%         & 33.42\%                       & 55.53\% ± 0.360  & 60.68\%         & 58.17\%         \\
          & 2                             & 75.33\% ± 0.272 & 72.13\%         & 84.11\%                       & 28.82\% ± 0.412 & 42.51\%         & 26.45\%                       & 52.08\% ± 0.342 & 57.32\%         & 55.28\%         \\
          & 3                             & 77.72\% ± 0.246 & 74.59\%         & 87.02\%                       & 31.98\% ± 0.411 & 45.07\%         & 28.64\%                       & 54.85\% ± 0.329 & 59.83\%         & 57.83\%         \\
\midrule
%\multirow{4}{*}{\textbf{t1 RGBof}}
  & 0                             & 77.80\% ± 0.220 & 73.80\%         & 87.82\%                       & 48.50\% ± 0.441 & 69.11\%         & 45.66\%                       & 63.15\% ± 0.331 & 71.46\%         & 66.74\%         \\
       \textbf{t1}~~~~ & 1                             & 80.15\% ± 0.202 & 76.55\%         & 88.87\%                       & 54.44\% ± 0.432 & 71.52\%         & 51.40\%                       & 67.30\% ± 0.317 & 74.04\%         & 70.14\%         \\
        \textbf{RGBof}  & 2                             & 74.17\% ± 0.230 & 68.21\%         & 87.98\%                       & 39.74\% ± 0.438 & 65.44\%         & 37.13\%                       & 56.96\% ± 0.334 & 66.83\%         & 62.56\%         \\
          & 3                             & 77.33\% ± 0.230 & 73.72\%         & 86.67\%                       & 49.70\% ± 0.437 & 69.10\%         & 46.50\%                       & 63.52\% ± 0.334 & 71.41\%         & 66.59\%         \\
\midrule
%\multirow{4}{*}{\textbf{t5 RGBof}}
  & 0                             & 76.39\% ± 0.254 & 72.65\%         & 85.77\%                       & 49.61\% ± 0.431 & 69.45\%         & 45.90\%                       & 63.00\% ± 0.343 & 71.05\%         & 65.84\%         \\
        \textbf{t5}~~~~  & 1                             & 75.81\% ± 0.266 & 72.14\%         & 84.15\%                       & 53.71\% ± 0.419 & 75.31\%         & 49.40\%                       & 64.76\% ± 0.343 & 73.73\%         & 66.78\%         \\
         \textbf{RGBof} & 2                             & 76.65\% ± 0.249 & 73.09\%         & 85.81\%                       & 51.71\% ± 0.434 & 71.14\%         & 48.36\%                       & 64.18\% ± 0.342 & 72.12\%         & 67.09\%         \\
          & 3                             & 77.55\% ± 0.245 & 73.82\%         & 87.02\%                       & 43.68\% ± 0.440 & 63.87\%         & 40.74\%                       & 60.62\% ± 0.343 & 68.85\%         & 63.88\%         \\
\midrule
%\multirow{4}{*}{\textbf{t1 XY}}
    & 0                             & 77.31\% ± 0.253 & 74.23\%         & 86.05\%                       & 49.95\% ± 0.444 & 66.71\%         & 47.06\%                       & 63.63\% ± 0.349 & 70.47\%         & 66.56\%         \\
         \textbf{t1}~ & 1                             & 77.46\% ± 0.243 & 73.90\%         & 87.27\%                       & 37.33\% ± 0.445 & 56.53\%         & 35.64\%                       & 57.40\% ± 0.344 & 65.22\%         & 61.46\%         \\
         \textbf{XY} & 2                             & 77.44\% ± 0.254 & 74.41\%         & 85.64\%                       & 52.91\% ± 0.433 & 70.33\%         & 49.46\%                       & 65.18\% ± 0.344 & 72.37\%         & 67.55\%         \\
          & 3                             & 76.20\% ± 0.255 & 72.38\%         & 86.22\%                       & 50.50\% ± 0.426 & 72.61\%         & 46.50\%                       & 63.35\% ± 0.341 & 72.50\%         & 66.36\%         \\
\midrule
%\multirow{4}{*}{\textbf{t5 XY}}
     & 0                             & 77.83\% ± 0.236 & 74.37\%         & 87.26\%                       & 46.42\% ± 0.437 & 66.69\%         & 43.20\%                       & 62.13\% ± 0.337 & 70.53\%         & 65.23\%         \\
         \textbf{t5}~ & 1                             & 79.70\% ± 0.224 & 76.53\%         & 88.18\%                       & 50.68\% ± 0.430 & 70.75\%         & 47.07\%                       & 65.19\% ± 0.327 & 73.64\%         & 67.63\%         \\
          \textbf{XY} & 2                             & 75.52\% ± 0.250 & 71.57\%         & 85.29\%                       & 50.37\% ± 0.426 & 71.05\%         & 46.44\%                       & 62.95\% ± 0.338 & 71.31\%         & 65.87\%         \\
          & 3                             & 78.03\% ± 0.225 & 74.60\%         & 88.41\%                       & 36.32\% ± 0.420 & 58.20\%         & 33.15\%                       & 57.18\% ± 0.323 & 66.40\%         & 60.78\%         \\
\midrule
%\multirow{4}{*}{\textbf{t1 PC}}
    & 0                             & 77.09\% ± 0.257 & 73.69\%         & 85.93\%                       & 46.06\% ± 0.451 & 63.74\%         & 43.63\%                       & 61.58\% ± 0.354 & 68.72\%         & 64.78\%         \\
         \textbf{t5}~ & 1                             & 77.81\% ± 0.246 & 73.78\%         & 87.90\%                       & 36.71\% ± 0.440 & 52.95\%         & 34.47\%                       & 57.26\% ± 0.343 & 63.37\%         & 61.19\%         \\
          \textbf{PC} & 2                             & 76.53\% ± 0.266 & 72.82\%         & 85.79\%                       & 42.95\% ± 0.450 & 66.04\%         & 40.85\%                       & 59.74\% ± 0.358 & 69.43\%         & 63.32\%         \\
          & 3                             & 76.33\% ± 0.249 & 72.72\%         & 84.70\%                       & 46.84\% ± 0.451 & 66.19\%         & 44.41\%                       & 61.59\% ± 0.350 & 69.46\%         & 64.56\%         \\
\midrule
%\multirow{4}{*}{\textbf{t5 PC}}
    & 0                             & 76.58\% ± 0.255 & 72.39\%         & 86.29\%                       & 44.95\% ± 0.434 & 66.53\%         & 41.47\%                       & 60.77\% ± 0.345 & 69.46\%         & 63.88\%         \\
         \textbf{t5}~  & 1                             & 74.77\% ± 0.279 & 71.22\%         & 83.69\%                       & 43.49\% ± 0.435 & 66.95\%         & 40.18\%                       & 59.13\% ± 0.357 & 69.09\%         & 61.94\%         \\
         \textbf{PC} & 2                             & 77.30\% ± 0.258 & 73.56\%         & 86.25\%                       & 45.89\% ± 0.446 & 64.58\%         & 43.09\%                       & 61.60\% ± 0.352 & 69.07\%         & 64.67\%         \\
          & 3                             & 77.41\% ± 0.234 & 72.87\%         & 87.85\%                       & 43.54\% ± 0.422 & 70.24\%         & 39.60\%                       & 60.48\% ± 0.328 & 71.56\%         & 63.73\%         \\
\bottomrule
\end{tabular}
\end{minipage}
\end{sideways}